\documentclass{article}

\PassOptionsToPackage{round,authoryear}{natbib}
 \usepackage[preprint]{neurips_2026}

\usepackage[normalem]{ulem}
\usepackage{xcolor}

\usepackage{cancel}

\usepackage[utf8]{inputenc} % allow utf-8 input
\usepackage[T1]{fontenc}    % use 8-bit T1 fonts
\usepackage{hyperref}       % hyperlinks
\usepackage{url}            % simple URL typesetting
\usepackage{booktabs}       % professional-quality tables
\usepackage{amsfonts}       % blackboard math symbols
\usepackage{nicefrac}       % compact symbols for 1/2, etc.
\usepackage{microtype}      % microtypography
\usepackage{xcolor}         % colors
\usepackage[pdftex]{graphicx}
\usepackage{amsmath}

\usepackage{algorithm}
\usepackage{algpseudocode}
\usepackage{mathtools}
\usepackage{amsthm}
\usepackage{subcaption}
\usepackage{caption}

\theoremstyle{definition}

\theoremstyle{remark}

\title{Dreaming Smoothly and Sample Efficiently with Gradient Penalized Latent Dynamics}
\author{
  Romil V.~Sonigra \\
  Department of Electrical and Computer Engineering\\
  Texas A\&M University\\
  College Station, TX 77843 \\
  \texttt{romils@tamu.edu} \\
  \And
  P. R. Kumar \\
  Department of Electrical and Computer Engineering\\
  Texas A\&M University\\
  College Station, TX 77843 \\
  \texttt{prk@tamu.edu} \\
}

\begin{document}

\maketitle

%Current final
% \begin{abstract}
% Model-based reinforcement learning improves sample efficiency by learning a world model. However, existing latent world models such as DreamerV3 do not explicitly enforce local smoothness in their learned transition dynamics, leaving a useful inductive bias for transition dynamics learning unexploited. We propose GPLD, a gradient-penalized latent dynamics regularizer for DreamerV3 that applies a row-wise Jacobian penalty to the posterior latent distribution to encourage locally smooth transition learning. We show that this penalty can be interpreted as the continuous-latent analogue of finite-difference smoothing of transition laws in discrete embedded-state MDPs, and estimate it efficiently using Hutchinson-style stochastic probes. Empirically, across DeepMind Control proprioceptive tasks, GPLD improves aggregate sample efficiency, with particularly strong gains on higher-complexity locomotion environments. On more challenging quadruped tasks, the benefits become more pronounced over longer training horizons. 
% % These results suggest that 
% Explicit local smoothness regularization is a simple and effective way to improve 
% latent world models for smooth continuous control environments.
% % ,while highlighting limitations in environments where discontinuities are more prominent.
% \end{abstract}
\begin{abstract}
Model-based reinforcement learning improves sample efficiency by learning a world model. However, existing latent world models such as DreamerV3 do not explicitly enforce local smoothness in their learned transition dynamics, leaving a useful inductive bias for transition dynamics learning unexploited. We propose GPLD, a gradient-penalized latent dynamics regularizer for DreamerV3 that applies a row-wise Jacobian penalty to the posterior latent distribution to encourage locally smooth transition learning. We show that this penalty can be interpreted as the continuous-latent analog of finite-difference smoothing of transition laws in discrete embedded-state MDPs, and estimate it efficiently using Hutchinson-style stochastic probes. Empirically, across DeepMind Control proprioceptive tasks, GPLD improves aggregate sample efficiency, with particularly strong gains on higher-complexity locomotion environments. On more challenging quadruped tasks, GPLD reaches high-return behavior earlier and exhibits more consistent late-stage learning over longer horizons. Explicit local smoothness regularization is a simple and effective way to improve latent world models for smooth continuous control environments. Code for GPLD is available at \href{https://github.com/romils9/gpld-mbrl}{\texttt{github.com/romils9/gpld-mbrl}}.
\end{abstract}

% % \include{sections/sec0_instructions}
%%%%%%%%%%%%%%%%%%%%%%%%%%%%%%%%%%%%%%%%%
% Start sections:
%%%%%%%%%%%%%%%%%%%%%%%%%%%%%%%%%%%%%%%%%
% \include{sections/sec1_intro}
%%%%%%%%%%%%%%%%%%%%%%%%%%%%%%%%%%%%%%%%%
\section{Introduction}
Model-based reinforcement learning (MBRL) improves sample efficiency by learning a predictive world model that can be used for planning and policy optimization~\citep{worldmodel_ha18}. DreamerV3 has demonstrated that latent world models can scale to diverse continuous-control tasks by learning compact recurrent state-space representations 
% \citep{hafner19a} 
and training policies through imagined rollouts~\citep{hafner2025dreamerv3}. Despite this success, standard latent world models do not explicitly exploit a basic structural prior of many continuous-control systems: nearby states induce similar short-horizon transition behavior. Encoding this prior directly in the learned latent dynamics can help the world model share information across nearby states, reducing the amount of interaction data needed to learn useful transition structure.

We study how to exploit this local smoothness prior through \emph{gradient-penalized latent dynamics} (GPLD), a differentiable regularizer for latent world-model dynamics.
This regularizer is motivated through a discrete-to-continuous smoothness argument. In a finite embedded-state MDP, local smoothness can be imposed by penalizing differences between transition laws at neighboring states. When the state representation is continuous and the transition model is differentiable, these normalized finite differences become directional derivatives of the learned dynamics. Averaging over local directions yields a Frobenius Jacobian penalty. This provides a principled route from neighborhood smoothing in tabular MDPs to the gradient penalty used by GPLD.

We implement GPLD in DreamerV3 and evaluate it on DeepMind Control tasks. Empirically, GPLD improves aggregate sample efficiency on proprioceptive tasks, with the largest gains attained on higher-complexity locomotion environments. On difficult quadruped tasks, the benefit becomes clearer over longer horizons, where GPLD reaches high-return behavior earlier and exhibits more consistent late-stage learning. 
Pixel-observation results show milder aggregate gains, suggesting that the effect of latent smoothness regularization is weaker when dynamics learning is coupled to high-dimensional visual encoding.

Our contributions are:
\begin{enumerate}
    \item A discrete-to-continuous justification showing how finite-difference smoothing of tabular transition laws leads to Frobenius Jacobian regularization in differentiable latent models.
    \item GPLD, a posterior Jacobian regularizer for DreamerV3 that explicitly encourages local smoothness in latent probability maps, yielding a normalized aggregate gain of \(17.7\%\) across the DeepMind Control proprioceptive benchmark.
    \item Evaluation of GPLD on DeepMind Control, showing a \(34.6\%\) normalized aggregate gain on higher-complexity proprioceptive locomotion tasks, clearer long-horizon benefits on difficult quadruped tasks, and ablations over the main design choices.
\end{enumerate}
Code for GPLD is available at
\href{https://github.com/romils9/gpld-mbrl}{\texttt{github.com/romils9/gpld-mbrl}}.

%%%%%%%%%%%%%%%%%%%%%%%%%%%%%%%%%%%%%%%%%
% \include{sections/sec3_related_work}
%%%%%%%%%%%%%%%%%%%%%%%%%%%%%%%%%%%%%%%%%
\section{Related Work}

\paragraph{Smoothness in reinforcement learning.}
Smoothness has long been recognized as important for generalization, stable prediction, and control in reinforcement learning~\citep{value_function_oscillations,munos2008finite,asadi18a,christmann2024benchmarkingsmoothnessreducinghighfrequency}. Unconstrained neural networks can produce high-frequency oscillations that degrade stability and generalization~\citep{value_function_oscillations,rosca_oscillations_smoothness_needed,christmann2024benchmarkingsmoothnessreducinghighfrequency}, and such errors can compound in multi-step value or dynamics prediction~\citep{Thrun_value_error_compounding,venkatraman_multi_step_prediction}. In model-based RL, \citet{asadi18a} show that Lipschitz constraints on learned transition models yield tighter multi-step estimation error bounds, directly linking transition smoothness to world-model accuracy. These works motivate smoothness as a useful inductive bias for learned dynamics, but do not provide a local Jacobian regularizer for stochastic latent world models.

\paragraph{Gradient penalties and control smoothness.}
Gradient penalties provide a differentiable way to encourage local smoothness and have been widely used in representation and generative modeling~\citep{gp_GANS,terjék2020adversariallipschitzregularization,goodfellow2015explainingharnessingadversarialexamples}. In reinforcement learning, related methods often regularize value functions, policies, or action trajectories. For example, \citet{model_free_Q_gp} apply gradient penalties to Q-functions in model-free RL, while \citet{chen2024learningsmoothhumanoidlocomotion} introduce Lipschitz-constrained policies as a differentiable alternative to task-specific reward-smoothing heuristics. Other control approaches encourage smooth behavior through penalties on mechanical energy or low-pass filtering~\citep{fu2021minimizingenergyconsumptionleads,peng2020lowpassfiltering}. In contrast, GPLD targets the learned world model itself: it applies a row-wise Jacobian penalty to the posterior latent probability map in DreamerV3, improving the latent dynamics used for inference and imagination.

\paragraph{Global versus local smoothness constraints.}
Global smoothness methods such as spectral normalization constrain the Lipschitz constant of neural networks by bounding layer-wise spectral norms~\citep{gogianu_spec_norm_in_rl,bjorck2022_drl_spec_norm}. However, global constraints can be overly restrictive because the global Lipschitz constant may arise from rare, unseen, or irrelevant regions of the input space~\citep{dherin2022neuralnetworkssimplesolutions}. GPLD instead imposes a local, data-dependent smoothness penalty on sampled latent inputs during world-model training. This distinguishes GPLD from global normalization methods and from policy-smoothing approaches: GPLD regularizes the posterior transition representation used by a model-based agent, while leaving the rest of the world model unconstrained except through the training objective.

%%%%%%%%%%%%%%%%%%%%%%%%%%%%%%%%%%%%%%%%%%%%%%%%%%%%%%%%
\section{A discrete-to-continuous justification for GPLD}\label{sec3:theory}
We justify GPLD by connecting a discrete local-smoothness prior on transition laws to the continuous Jacobian penalty used in our latent world-model setting. We begin with a finite embedded-state MDP, where smoothness can be imposed by penalizing finite differences of neighboring transition distributions. We then consider a differentiable transition model over continuous state representations and show that these finite differences become directional Jacobian norms, whose isotropic average yields the Frobenius Jacobian norm. This produces the row-wise posterior regularizer used in GPLD.

\subsection{Discrete local smoothness prior}

We begin with a finite state and finite action MDP. Let $\mathcal{S}$ denote the state space, $\mathcal{A}$ the action space, and let $e:\mathcal{S}\to\mathbb{R}^d$ be an embedding of states into a metric space. For each action $a\in\mathcal{A}$, let $P(\cdot\mid s,a)\in\Delta^{|\mathcal{S}|-1}$ denote the transition probability vector, and define the $\varepsilon$-neighborhood of $s$ by
\begin{equation*}
\mathcal{N}_\varepsilon(s) := \{ k \in \mathcal{S} : \|e(k)-e(s)\|_2 \le \varepsilon \}.
% \label{eq:eps_neighborhood}
\end{equation*}

Given a dataset of transitions $\mathcal{D}=\{(s,a,s')\}_{t=1}^T$, let $N(s,a,s')$ denote the number of observed transitions from $(s,a)$ to $s'$. The standard maximum-likelihood estimator minimizes the negative log-likelihood
\begin{equation*}
    \mathcal{L}_{\mathrm{MLE}}(P;\mathcal{D})
=
-\sum_{s,a,s'} N(s,a,s') \log P(s' \mid s,a).
\end{equation*}

A natural structural prior is that nearby states should induce similar transition laws. We encode this by coupling neighboring transition vectors through a finite-difference regularizer:
\begin{equation}
\mathcal{L}_{\mathrm{FD}}(P)
=
\sum_{a}\sum_{s}\sum_{k\in\mathcal{N}_\varepsilon(s)}\sum_{s'}
\left(
\frac{P(s'\mid k,a)-P(s'\mid s,a)}{\|e(k)-e(s)\|_2}
\right)^2.
\label{eq:fd_regularizer}
\end{equation}
This yields the regularized objective
\begin{equation*}
\min_{P}\;
\mathcal{L}_{\mathrm{MLE}}(P;\mathcal{D})
+
\lambda \mathcal{L}_{\mathrm{FD}}(P).
% \label{eq:regularized_tabular_mle}
\end{equation*}

% Equation~\eqref{eq:regularized_tabular_mle}
This is the discrete precursor of GPLD: it penalizes normalized local differences of transition laws across neighboring states and therefore explicitly favors locally smooth transition estimates.

\subsection{Continuous-state limit}
The discrete finite-difference regularizer in \eqref{eq:fd_regularizer} 
leads to a Jacobian penalty when the transition law is represented by a differentiable function over continuous state embeddings.
Consider a neighboring state \(k \in \mathcal{N}_\varepsilon(s)\). In the embedding space, its displacement from \(s\) can be written as
\begin{equation}
e(k)-e(s)=hu,
\label{eq:local_displacement}
\end{equation}
where \(h=\|e(k)-e(s)\|_2\) and \(u\in\mathbb{R}^d\) is a unit vector. Substituting \eqref{eq:local_displacement} into a summand of \eqref{eq:fd_regularizer} yields
\begin{equation*}
\left\|
\frac{P(\cdot \mid k,a)-P(\cdot \mid s,a)}{\|e(k)-e(s)\|_2}
\right\|_2^2
\;\leadsto\;
\left\|
\frac{f_\theta(x+hu,a)-f_\theta(x,a)}{h}
\right\|_2^2,
% \label{eq:fd_to_continuous}
\end{equation*}
where \(x=e(s)\), and \(f_\theta\) denotes a differentiable continuous-state transition model. Thus, the discrete regularizer naturally gives rise to normalized local finite differences of \(f_\theta\).

The small-neighborhood limit \(h\to 0\) then converts these finite differences into directional derivatives.

Let \(f_\theta:\mathbb{R}^d \to \mathbb{R}^m\) be locally Fr\'echet differentiable at \(x\in\mathbb{R}^d\). Then for any unit vector \(u\in\mathbb{R}^d\),
\begin{equation}
\lim_{h\to 0}
\left\|
\frac{f_\theta(x+hu)-f_\theta(x)}{h}
\right\|_2^2
=
\|J_{f_\theta}(x)u\|_2^2.
\label{eq:directional_limit_clean}
\end{equation}

% Proposition~\ref{prop:directional_limit} 
This shows that each normalized finite-difference term converges to $\|J_{f_\theta}(x)u\|_2^2$.
% the squared $\ell_2$-norm of the directional derivative of $f_\theta$ at $x$ along the unit direction $u$. 
Since the discrete regularizer aggregates neighboring states around each reference state, its small-neighborhood continuous limit induces an average over all unit directions around $x$. 

To see how isotropic averaging yields Frobenius Jacobian energy, let \(u\) be uniformly distributed on the unit sphere \(\mathbb{S}^{d-1}\). Then
\begin{equation}
\mathbb{E}_{u}\!\left[\|J_{f_\theta}(x)u\|_2^2\right]
=
\frac{1}{d}\|J_{f_\theta}(x)\|_F^2.
\label{eq:isotropic_average_clean}
\end{equation}

Thus, under a local isotropy assumption, the continuous counterpart of the discrete finite-difference regularizer is, up to a constant factor, a squared Frobenius Jacobian penalty. This establishes the theoretical bridge from discrete neighborhood smoothing to the continuous Jacobian regularization used by GPLD. In the next section, we instantiate this penalty in DreamerV3 by applying it row-wise to the posterior latent probability map. Proofs of \eqref{eq:directional_limit_clean} and \eqref{eq:isotropic_average_clean} are deferred to Appendix~\ref{appA:theory_proofs}.

%%%%%%%%%%%%%%%%%%%%%%%%%%%%%%%%%%%%%%%%%
% \include{sections/sec4_algorithm}
%%%%%%%%%%%%%%%%%%%%%%%%%%%%%%%%%%%%%%%%%
\section{Gradient-Penalized Latent Dynamics in DreamerV3}
\label{sec:gpld_dreamerv3}

We now instantiate the Frobenius Jacobian penalty motivated in Section~\ref{sec3:theory} within DreamerV3. GPLD regularizes the posterior latent probability map rather than the raw environment transition map. This choice matches the object used for latent-state inference during world-model learning and yields a direct implementation through reverse-mode automatic differentiation.

DreamerV3 trains a recurrent state-space world model with deterministic latent state \(h_t\), stochastic latent state \(z_t\), encoder output \(e_t\), and action \(a_t\), all at time \(t\). Its world-model objective can be written as
\[
\mathcal{L}_{\mathrm{Dv3}}
:=
\mathbb{E}_{q_\phi}
\left[
\sum_{t=1}^{T}
\left(
\beta_{\mathrm{pred}}\mathcal{L}_{\mathrm{pred},t}
+
\beta_{\mathrm{dyn}}\mathcal{L}_{\mathrm{dyn},t}
+
\beta_{\mathrm{rep}}\mathcal{L}_{\mathrm{rep},t}
\right)
\right],
\]
where \(\mathcal{L}_{\mathrm{pred},t}\) combines reconstruction, reward, and continuation prediction losses at time \(t\). The dynamics and representation KL terms are
\[
\mathcal{L}_{\mathrm{dyn},t}
=
\max\left(
1,
D_{\mathrm{KL}}\!\left(
\mathrm{sg}\!\left[q_\phi(z_t\mid h_t,e_t)\right]
\,\|\, 
p_\phi(z_t\mid h_t)
\right)
\right),
\]
and
\[
\mathcal{L}_{\mathrm{rep},t}
=
\max\left(
1,
D_{\mathrm{KL}}\!\left(
q_\phi(z_t\mid h_t,e_t)
\,\|\,
\mathrm{sg}\!\left[p_\phi(z_t\mid h_t)\right]
\right)
\right),
\]
where \(\mathrm{sg}[\cdot]\) denotes the stop-gradient operator, which prevents gradients from flowing through its argument.

Let \(u_t := [h_t,e_t]\) denote the posterior input. After row-wise softmax, the posterior distribution \(q_\phi(z_t\mid h_t,e_t)\) is represented as a \(K\times C\) probability table
\[
q_\phi(u_t) \in [0,1]^{K\times C},
\]
where \(K\) is the number of categorical variables and \(C\) is the number of classes per variable. 
% Each row of this table sums to one and therefore defines a categorical distribution. 
We refer to the map \(u_t \mapsto q_\phi(u_t)\) as the posterior latent probability map. Let \(q_\phi^{(i)}(u_t)\in\Delta^{C-1}\) denote the \(i\)-th row, where \(\Delta^{C-1}\) is the probability simplex over \(C\) classes.

GPLD penalizes the average row-wise Frobenius Jacobian norm of this posterior latent probability map with respect to \(u_t\):
\begin{equation}
\mathcal{R}_{\mathrm{GPLD}}(u_t)
=
\frac{1}{K}
\sum_{i=1}^{K}
\left\|
J_{q_\phi^{(i)}}(u_t)
\right\|_{F}^{2}.
\label{eq:gpld_rowwise_penalty}
\end{equation}
The resulting world-model objective is
\begin{equation*}
\mathcal{L}_{\mathrm{GPLD}}
:=
\mathbb{E}_{q_\phi}
\left[
\sum_{t=1}^{T}
\left(
\beta_{\mathrm{pred}}\mathcal{L}_{\mathrm{pred},t}
+
\beta_{\mathrm{dyn}}\mathcal{L}_{\mathrm{dyn},t}
+
\beta_{\mathrm{rep}}\mathcal{L}_{\mathrm{rep},t}
+
\lambda^{\mathrm{post}}_t \mathcal{R}_{\mathrm{GPLD}}(u_t)
\right)
\right].
% \label{eq:gpld_world_model_loss}
\end{equation*}

Computing the full Jacobian in \eqref{eq:gpld_rowwise_penalty} explicitly is expensive. We therefore estimate each row-wise Frobenius term using a Hutchinson-style output-space probe. For each posterior row \(i\), sample \(\epsilon_i\sim\mathrm{Rad}^{C}\). Then
\begin{equation*}
g_i(u_t)
:=
\nabla_{u_t}
\left\langle
\epsilon_i,
q_\phi^{(i)}(u_t)
\right\rangle
=
J_{q_\phi^{(i)}}(u_t)^\top \epsilon_i.
% \label{eq:gpld_vjp}
\end{equation*}
Since \(\mathbb{E}[\epsilon_i\epsilon_i^\top]=I_C\), we have
\begin{equation*}
\mathbb{E}_{\epsilon_i}
\left[
\left\|
J_{q_\phi^{(i)}}(u_t)^\top\epsilon_i
\right\|_2^2
\right]
=
\left\|
J_{q_\phi^{(i)}}(u_t)
\right\|_F^2.
% \label{eq:gpld_hutchinson_identity}
\end{equation*}
Thus, the implemented estimator is
\begin{equation*}
\widehat{\mathcal{R}}_{\mathrm{GPLD}}(u_t)
=
\frac{1}{K}
\sum_{i=1}^{K}
\left\|
J_{q_\phi^{(i)}}(u_t)^\top \epsilon_i
\right\|_2^2.
% \label{eq:gpld_hutchinson_estimator}
\end{equation*}
Algorithm~\ref{alg:gp_post} summarizes the minibatch implementation.

\begin{algorithm}[t]
\caption{Gradient penalty on the posterior latent probability map using Hutchinson's estimator}
\label{alg:gp_post}
\begin{algorithmic}[1]
\State \textbf{Input:} batch of latent states $h$, encoded observations $e$, posterior model $q(h,e)$, penalty coefficient $\lambda_T$, sampling fraction $\rho$
\State \textbf{Output:} posterior gradient penalty $\lambda_T \, L_{\mathrm{gp}}$

\State $B \gets$ total batch size
\State $N \gets \lfloor \rho B \rfloor$ \hfill // number of sampled states
\State sample index set $I \subset \{1,\dots,B\}$ with $|I| = N$
\State $\mathcal{L}_{\mathrm{GPLD}} \gets 0$

\For{each index $k \in I$}
    \State extract $h_k, e_k$ from the batch
    \State $x_k \gets [h_k, e_k]$
    \State $\mathbf{q}_k \gets q(h_k, e_k) \in [0,1]^{32 \times 16}$ \hfill // row-wise probabilities
    \For{$i = 1$ to $K$}
        \State sample Rademacher noise $\epsilon_{k,i} \sim \{-1,+1\}^{C}$
        \State $s_{k,i} \gets \langle \epsilon_{k,i}, \mathbf{q}_{k,i} \rangle$
        \State $g_{k,i} \gets \nabla_{x_k} s_{k,i}$
        \State $L_{\mathrm{gp}} \gets L_{\mathrm{gp}} + \|g_{k,i}\|_2^2$
    \EndFor
\EndFor
\State $L_{\mathrm{gp}} \gets \frac{1}{K\cdot N} L_{\mathrm{gp}}$
\State \Return $\lambda_T \, L_{\mathrm{gp}}$
\end{algorithmic}
\end{algorithm}

\subsection{Why regularize only the posterior?}
\label{sec:posterior_only}

GPLD regularizes the posterior latent probability map rather than both the posterior and prior. This targets the observation-conditioned distribution \(q_\phi(z_t\mid h_t,e_t)\), which is used to infer latent states from real trajectories during world-model training. Since the posterior depends directly on the encoded observation \(e_t\), regularizing this map constrains the latent representation learned from environment data.

The prior \(p_\phi(z_t\mid h_t)\) is not directly penalized by GPLD, but it is trained against the posterior through the dynamics KL term \(\mathcal{L}_{\mathrm{dyn},t}\). Thus, posterior-only regularization directly smooths the observation-conditioned latent map while still influencing the target that the prior learns to predict. Ablations in Section~\ref{sec5.4:ablations} show that this choice provides the best performance--cost trade-off among posterior-only, prior-only, and joint regularization.

\subsection{Time-decayed gradient penalty coefficient}
\label{sec:lambda_decay}

A fixed smoothness penalty can be useful early in training, when the world model is poorly estimated, but may become overly restrictive later as the model fit improves. We therefore decay the posterior penalty coefficient over training. Our default schedule uses square-root decay with a minimum threshold:
\[
\lambda^{\mathrm{post}}_T
=
\max\left(
\frac{\lambda^{\mathrm{post}}_0}{\sqrt{1 + T_{\mathrm{updates}}/c}},
\lambda_{\min}
\right),
\]
where \(T_{\mathrm{updates}}\) is the cumulative number of optimizer updates and \(c>0\) is a decay-scale constant. The square-root form is motivated by the finite-state intuition that explicit smoothing should matter most when transition estimates are data-limited, and should weaken as more data are collected. Algorithm~\ref{alg:lambda_decay} summarizes the schedule.

\begin{algorithm}[t]
\caption{Square-root decay of the gradient penalty coefficient}
\label{alg:lambda_decay}
\begin{algorithmic}[1]
\State \textbf{Input:} initial coefficient $\lambda_0$, current update step $T_{\mathrm{updates}}$, decay scale \(c\), minimum coefficient \(\lambda_{\min}\)
\State \textbf{Output:} decayed coefficient $\lambda_T$

\State $s \gets 1 + T_{\mathrm{updates}} / c$
\State $\lambda_T \gets \max\left(\lambda_0 / \sqrt{s}, \lambda_{\min}\right)$
\State \Return $\lambda_T$
\end{algorithmic}
\end{algorithm}

%%%%%%%%%%%%%%%%%%%%%%%%%%%%%%%%%%%%%%%%%
% \include{sections/sec5_experiments}
%%%%%%%%%%%%%%%%%%%%%%%%%%%%%%%%%%%%%%%%%
\section{Experiments}
\label{sec:experiments}

We evaluate GPLD on DeepMind Control (DMC) tasks~\citep{tunyasuvunakool2020} to test whether local smoothness regularization improves latent world-model learning. Our main evaluation uses proprioceptive observations, where the agent receives low-dimensional physical state features. We additionally evaluate pixel observations, where the agent receives image frames and must learn visual representations jointly with latent dynamics.

\subsection{Experimentation}
\label{sec5.1:experimentation}

\begin{figure}[t]
  \centering

  {\small\textbf{Cheetah run}} \\[0.1em]
  \includegraphics[width=0.86\linewidth]{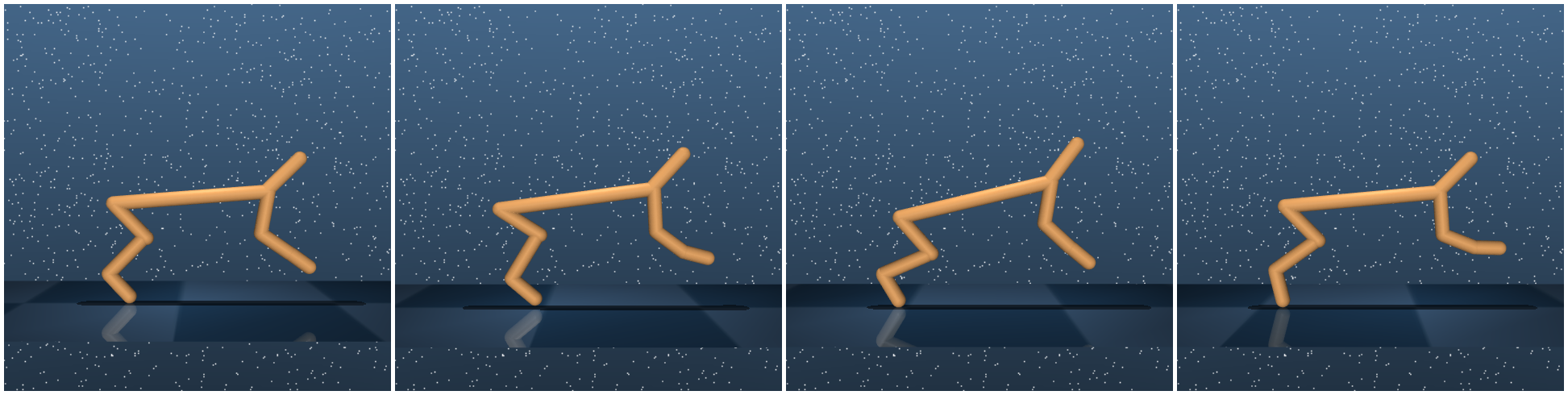}

  \vspace{0.5em}

  {\small\textbf{Walker walk}} \\[0.1em]
  \includegraphics[width=0.86\linewidth]{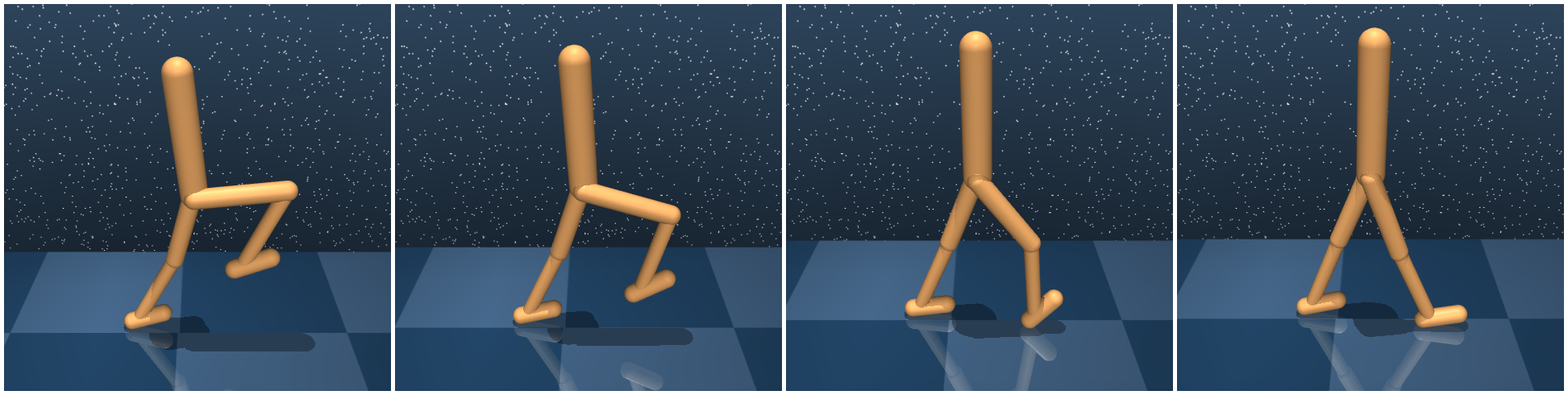}

  \caption{Representative frames from DMC locomotion tasks. Many proprioceptive locomotion environments evolve through smooth body motions over short time intervals, making them a natural testbed for evaluating local smoothness regularization in latent dynamics.}
  \label{fig:smoothness_intuition}
\end{figure}

DMC locomotion tasks provide a natural testbed for GPLD because many of them evolve through locally smooth body motions over short time intervals, as qualitatively illustrated in Fig.~\ref{fig:smoothness_intuition}. We compare GPLD-DreamerV3 against DreamerV3 using aggregate learning curves, representative task-level curves, and ablations over the main design choices. Experimental details, including hyperparameters, seed counts, evaluation protocol, and aggregate normalization, are provided in Appendix~\ref{app:experimental_details}.

\subsection{DMC Proprioceptive Tasks}
\label{sec5.2:dmc_proprio}
\begin{figure}[ht]
  \vskip 0.2in
  \centering
  \begin{minipage}[t]{0.49\linewidth}
    \centering
    \includegraphics[width=\linewidth]{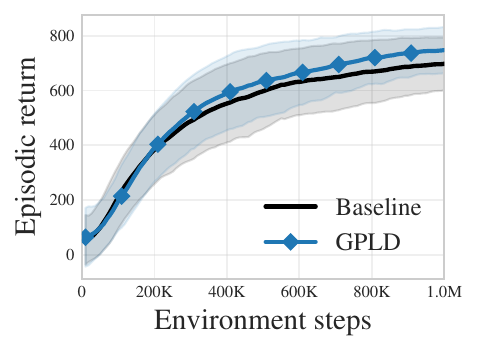}\\[-2pt]
    {\footnotesize (a) Raw Aggregate Mean Score}
  \end{minipage}\hfill
  \begin{minipage}[t]{0.49\linewidth}
    \centering
    \includegraphics[width=\linewidth]{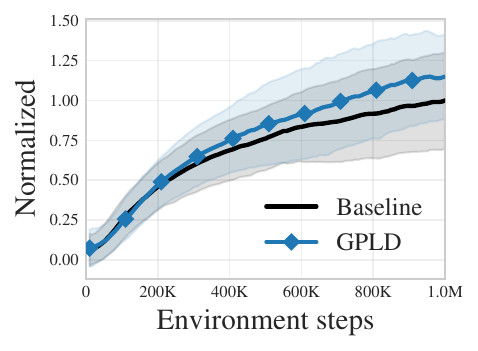}\\[-2pt]
    {\footnotesize (b) Normalized Aggregate Mean Score}
  \end{minipage}
  \caption{Aggregate proprioceptive-control performance on DMC tasks. GPLD-DreamerV3 improves sample efficiency over DreamerV3 in both raw aggregate return and normalized aggregate return.}
\label{fig:proprio_aggregate}
\end{figure}

GPLD provides the strongest evidence for its effectiveness in proprioceptive control, where the world model learns from low-dimensional physical state features and the local-smoothness prior acts directly on latent dynamics learning. Across the proprioceptive DMC benchmark, GPLD-DreamerV3 improves sample efficiency over DreamerV3, as shown in Fig.~\ref{fig:proprio_aggregate}. Importantly, these aggregate gains are obtained using a single GPLD configuration across tasks, without per-environment tuning. The improvement appears in both raw aggregate return and normalized aggregate return, indicating that GPLD does not merely improve a single high-return environment, but produces broader gains across the benchmark.

The largest gains appear on tasks that demand coordinated locomotion and richer transition modeling. On the higher-complexity proprioceptive locomotion subset in Table~\ref{tab:complex_proprio_1m}, GPLD improves the normalized aggregate score by \(34.6\%\) at 1M environment steps. 
The improvement is especially large on hopper-hop and walker-run, while gains are smaller on tasks where DreamerV3 already approaches high return. 
The 1M-step comparison across all proprioceptive tasks is reported in Appendix~\ref{app:proprio_2M},  Table~\ref{tab:app_proprio_1m_all_tasks}, where GPLD improves the normalized aggregate mean by \(17.7\%\).
Individual 2M-step learning curves for the proprioceptive suite are shown in Appendix~\ref{app:proprio_2M}, Fig.~\ref{fig:proprio_2M_grid}.

%%%%%%%%%%%%%%%%%%%%%%%%%%%%%%%%%%%%%%%%%%%%%%%%%%%%%%%%%%%%%%
%%%%%%%%%%%%%%%%%%%%%%%%%%%%%%%%%%%%%%%%%%%%%%%%%%%%%%%%%%%%%%

\begin{table}[t]
 \caption{Performance at 1M environment steps on higher-complexity proprioceptive DMC tasks. Individual task scores are raw episodic returns. For the aggregate summary reported in the text, scores are normalized by the final DreamerV3 baseline score for each task before aggregation.}
  \label{tab:complex_proprio_1m}
  \centering
  \begin{tabular}{lccc}
    \toprule
    Task & DreamerV3 & GPLD-DreamerV3 & Relative gain \\
    \midrule
    cheetah-run   & 672.6 & 762.8 & \(+13.4\%\) \\
    hopper-hop    & 81.8 & 197.4 & \(+141.3\%\) \\
    hopper-stand  & 683.6 & 730.5 & \(+6.9\%\) \\
    walker-run    & 458.5 & 627.1 & \(+36.8\%\) \\
    walker-stand  & 942.8 & 957.8 & \(+1.6\%\) \\
    walker-walk   & 878.3 & 946.2 & \(+7.7\%\) \\
    % \midrule
    % Aggregate     & 619.6 & 703.6 & \(+13.6\%\) \\
    % Normalized Aggregate & 100 & 133.8 & \(+34.6\%\) \\
    \bottomrule
  \end{tabular}
\end{table}

\begin{figure}[ht]
  \vskip 0.2in
  \centering
  \begin{minipage}[t]{0.49\linewidth}
    \centering
    \includegraphics[width=\linewidth]{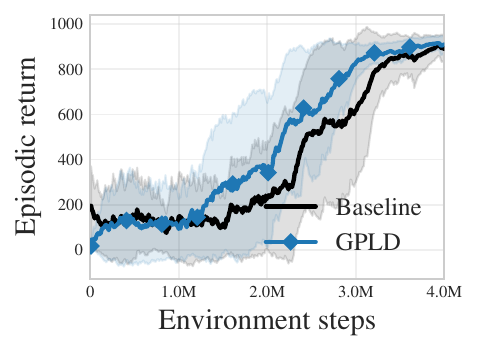}\\[-2pt]
    {\footnotesize (a) Quadruped run}
  \end{minipage}\hfill
  \begin{minipage}[t]{0.49\linewidth}
    \centering
    \includegraphics[width=\linewidth]{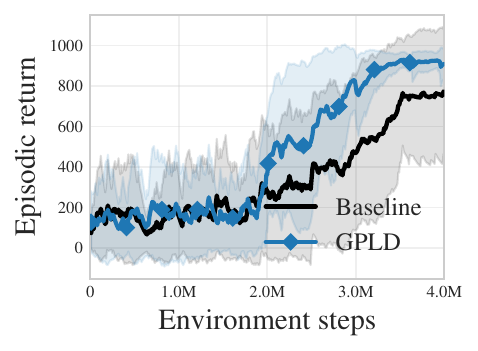}\\[-2pt]
    {\footnotesize (b) Quadruped Walk}
  \end{minipage}
  \caption{Long-horizon quadruped performance. Mean episodic return is reported across seeds, with shaded regions indicating one standard deviation around the mean. On quadruped-run and quadruped-walk, GPLD-DreamerV3 reaches high-return behavior earlier and maintains stronger late-stage performance than DreamerV3.}
\label{fig:quad_long_horizon}
\end{figure}

Difficult quadruped tasks reveal a complementary long-horizon effect. Unlike the 1M-step locomotion subset, quadruped-run and quadruped-walk remain noisy early in training for both methods, suggesting that reliable latent dynamics for these environments require substantially more interaction. Over 4M environment steps, however, GPLD-DreamerV3 reaches the high-return regime earlier and maintains stronger late-stage performance than DreamerV3, as shown in Fig.~\ref{fig:quad_long_horizon}. This indicates that the benefit of local smoothness regularization is not limited to early learning: on harder locomotion tasks, it can also improve the quality and consistency of learning over longer horizons.

%%%%%%%%%%%%%%%%%%%%%%%%%%%%%%%%%%%%%%%%%%%%%%%%%%%%%%%%%%%%

%%%%%%%%%%%%%%%%%%%%%%%%%%%%%%%%%%%%%%%%%
% \include{sections/sec5_experiments}
%%%%%%%%%%%%%%%%%%%%%%%%%%%%%%%%%%%%%%%%%
\subsection{DMC Pixel Tasks}
\label{sec5.3:dmc_pixel}

\begin{figure}[]
  \vskip 0.2in
  \centering
  \begin{minipage}[t]{0.24\columnwidth}
    \centering
    \includegraphics[width=\linewidth]{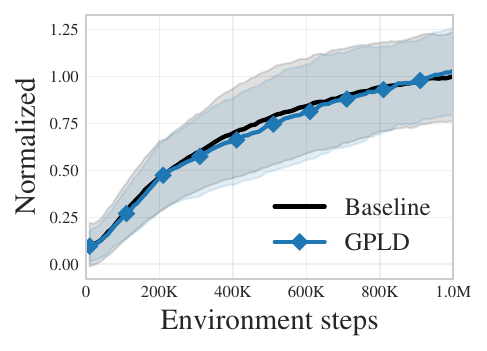}
    {\footnotesize (a) Normalized aggregate}
  \end{minipage}\hfill
  \begin{minipage}[t]{0.24\columnwidth}
    \centering
    \includegraphics[width=\linewidth]{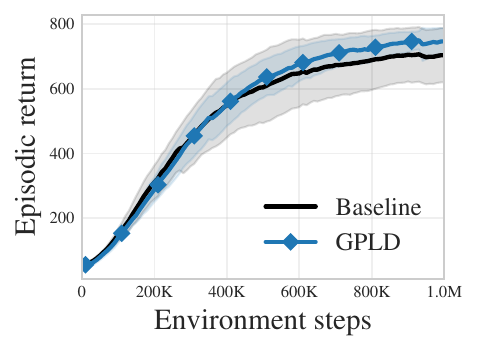}
    {\footnotesize (b)Walker-run}
  \end{minipage}
  \begin{minipage}[t]{0.24\columnwidth}
    \centering
    \includegraphics[width=\linewidth]{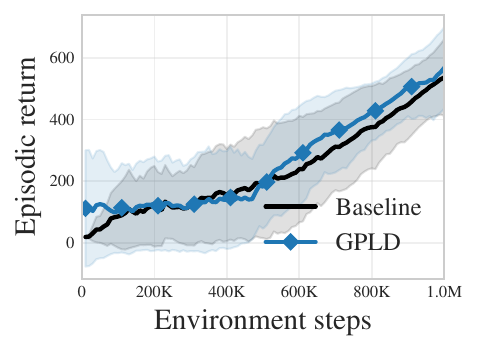}
    {\footnotesize (c) Quadruped-run}
  \end{minipage}\hfill
  \begin{minipage}[t]{0.24\columnwidth}
    \centering
    \includegraphics[width=\linewidth]{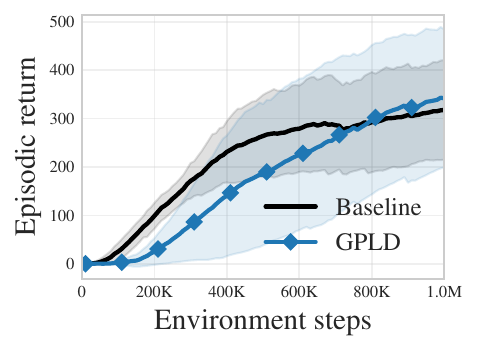}
    {\footnotesize (d)Hopper-hop}
  \end{minipage}
  \caption{Pixel-observation DMC results. With image-frame observations, GPLD-DreamerV3 remains competitive in normalized aggregate performance but shows smaller gains than in proprioceptive control. Representative curves show clearer late-stage improvements on walker-run and quadruped-run, while hopper-hop exhibits slower early learning and a smaller late-stage advantage.}
\label{fig:pixel_results}
\end{figure}

We also evaluate GPLD with pixel observations, where the agent receives image frames rather than low-dimensional proprioceptive state features. This setting is harder because the world model must learn visual encodings and latent dynamics jointly. As shown in Fig.~\ref{fig:pixel_results}, GPLD-DreamerV3 remains competitive with DreamerV3 in the normalized aggregate, but the gain is smaller than in proprioceptive control.

The task-level curves help explain this weaker aggregate effect. GPLD shows clearer late-stage gains on walker-run and quadruped-run, but hopper-hop is mixed: it learns more slowly early in training and catches up later. Thus, pixel results support a more cautious conclusion than the proprioceptive results. 
GPLD can still help under visual observations, but its effect is less direct when the same world model must simultaneously learn visual representations and smooth latent dynamics. 
Appendix~\ref{app:vision_grid} provides individual pixel-observation learning curves for all evaluated tasks, while 
Appendix~\ref{app:vision_warmstart} further studies an encoder-decoder warm-start diagnostic to separate the role of visual representation learning from latent-dynamics regularization.

\subsection{Ablations}
\label{sec5.4:ablations}
\begin{figure}[ht]
  \vskip 0.2in
  \centering
  \begin{minipage}[t]{0.24\linewidth}
    \centering
    \includegraphics[width=\linewidth]{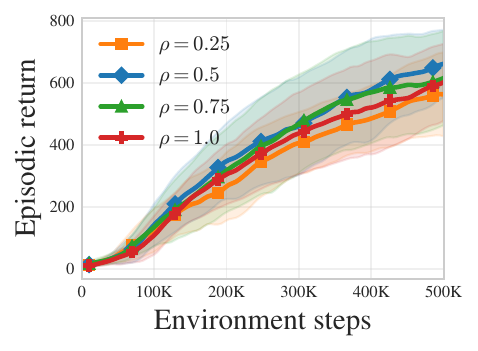}\\[-2pt]
    {\footnotesize (a) Vary $\rho$ at $\lambda_0^{\mathrm{post}}=0.5$}
  \end{minipage}\hfill
  \begin{minipage}[t]{0.24\linewidth}
    \centering
    \includegraphics[width=\linewidth]{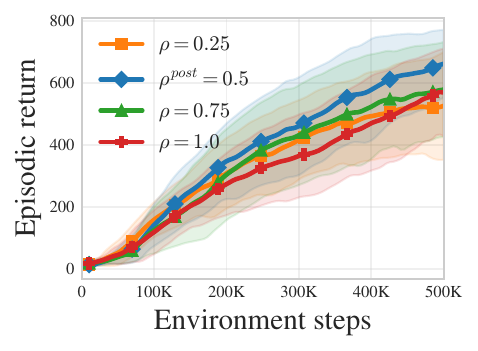}\\[-2pt]
    {\footnotesize (b) Vary $\rho$ with $\lambda_0^{\mathrm{prior}}=0.5$}
  \end{minipage}
  \begin{minipage}[t]{0.24\linewidth}
    \centering
    \includegraphics[width=\linewidth]{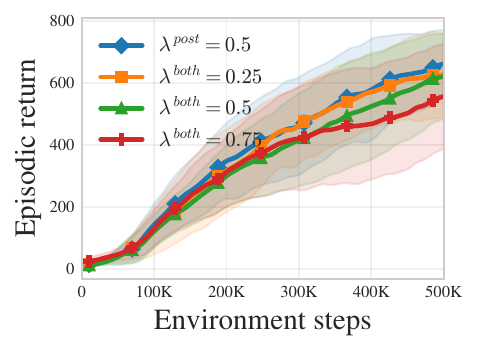}\\[-2pt]
    {\footnotesize (c) Vary $\lambda_0^{\mathrm{post}}$ and $\lambda_0^{\mathrm{prior}}$}
  \end{minipage}\hfill
  \begin{minipage}[t]{0.24\linewidth}
    \centering
    \includegraphics[width=\linewidth]{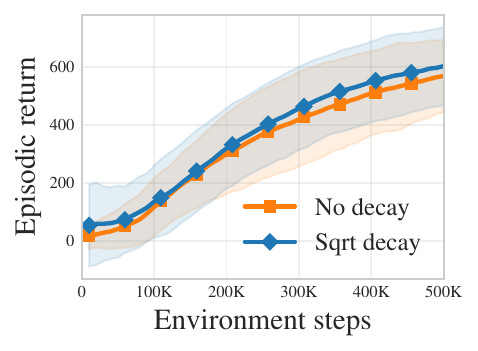}\\[-2pt]
    {\footnotesize (d) (d) Sqrt decay vs no decay}
  \end{minipage}
  \caption{GPLD ablations over sampling fraction, posterior/prior regularization, and penalty scheduling. The results support using GPLD as a targeted posterior regularizer with time-decayed smoothness rather than a broad or fixed smoothness constraint.}
  \label{fig4:ablations_proprio}
\end{figure}

The ablations in Fig.~\ref{fig4:ablations_proprio} show that GPLD works best as a targeted posterior regularizer rather than as a broad smoothness constraint on all latent distributions. Posterior-only regularization performs competitively with or better than prior-only and joint prior--posterior regularization, while avoiding the extra derivative cost of explicitly penalizing both distributions. This supports the design choice in Section~\ref{sec:posterior_only}: GPLD directly regularizes the observation-conditioned latent map, while the prior remains coupled through the DreamerV3 dynamics loss.

The sampling fraction \(\rho\) controls the fraction of batch states on which the gradient penalty is evaluated, while the penalty coefficient controls the strength of the smoothness penalty on those states. Increasing \(\rho\) applies GPLD to more batch states, but the gains do not increase monotonically enough to justify always using the largest value. The decay ablation further shows that maintaining a strong penalty throughout training is less effective than square-root decay, supporting the schedule in Section~\ref{sec:lambda_decay}. Overall, the ablations indicate that GPLD is most effective when applied selectively to the posterior with moderate, time-decayed regularization.

\subsection{Sensitivity diagnostics and computational cost}
\label{sec5.5:mechanism_cost}

We include two additional diagnostics to check whether GPLD behaves as intended. To assess the effect of the gradient penalty, we measure how much the learned posterior and prior distributions change under small perturbations of their inputs. A smoother latent map should change less under such perturbations. GPLD reduces this measured local sensitivity on representative locomotion tasks, supporting the intended effect of the regularizer. Full details and plots are provided in Appendix~\ref{app:local_sensitivity}.

To assess computational overhead, we measure wall-clock runtime relative to DreamerV3. The additional cost comes from estimating the gradient penalty through vector-Jacobian products, and scales with the number of batch states selected by the sampling fraction \(\rho\). Timing results are reported in Appendix~\ref{app:wall_clock}.

%%%%%%%%%%%%%%%%%%%%%%%%%%%%%%%%%%%%%%%%%
% \include{sections/sec7_conclusion}
%%%%%%%%%%%%%%%%%%%%%%%%%%%%%%%%%%%%%%%%%
\section{Conclusion}
\label{sec:conclusion}
We have introduced GPLD, a local smoothness regularizer for latent world models, and instantiated it in DreamerV3 through a row-wise Jacobian penalty on the posterior latent probability map. This penalty is motivated by a discrete-to-continuous argument showing how finite-difference smoothing of neighboring transition laws leads to Frobenius Jacobian regularization in differentiable latent models. 
Empirically, GPLD improves aggregate sample efficiency on proprioceptive DMC tasks, with the strongest gains on higher-complexity locomotion environments. The quadruped results further show that local smoothness regularization improves the quality and consistency of learning over longer horizons.
These gains are obtained with a single GPLD configuration across tasks, without per-environment tuning.

There are some important limitations. GPLD is not uniformly beneficial across all environments, and its gains are weaker in pixel-observation settings, where latent dynamics learning is coupled to high-dimensional visual encoding. More broadly, GPLD encodes a local smoothness prior, so it may be less appropriate in environments where discontinuities dominate the relevant dynamics.

\section*{Impact Statement}
This work aims to improve the sample efficiency of model-based reinforcement learning by regularizing learned latent dynamics. More sample-efficient world models can reduce the number of interactions needed to train continuous-control agents, potentially lowering simulation cost and making model-based RL more practical. As with other reinforcement-learning methods, downstream impact depends on the systems in which the trained agents are deployed. If such methods are later used in physical-control systems, deployment would require the same safety validation expected for other reinforcement-learning agents. Our experiments are limited to simulated DeepMind Control tasks and do not involve deployed robotic systems or human subjects.

\section*{Acknowledgements}
Portions of this research were conducted with the advanced computing resources provided by Texas A\&M High Performance Research Computing. This material is based upon work partially supported by the U.S. Army Contracting Command under Contract Numbers W911NF2120064, W911NF2520046, and W911NF2210151, the Office of Naval Research under Contracts N000142412615, and N00014-21-1-2385, and the National Science Foundation under Contract Numbers CNS-2328395 and CMMI-2038625. The views expressed herein and conclusions contained in this document are those of the authors and should not be interpreted as representing the views or official policies, either expressed or implied, of the U.S. Army, Navy, or the United States Government. The U.S. Government is authorized to reproduce and distribute reprints for Government purposes notwithstanding any copyright notation herein.

%%%%%%%%%%%%%%%%%%%%%%%%%%%%%%%%%%%%%%%%%%%%%%%%%%%%%%%%%%%%
% Bibliography-
\bibliographystyle{plainnat}
\bibliography{references}
%%%%%%%%%%%%%%%%%%%%%%%%%%%%%%%%%%%%%%%%%%%%%%%%%%%%%%%%%%%%

\appendix

\section{Proofs for the discrete-to-continuous justification of GPLD}\label{appA:theory_proofs}
% \subsection{Proof of Finite Differences converging to directional Jacobian sensitivities \eqref{eq:directional_limit_clean}}
% \label{app:proof_prop31}
\begin{enumerate}
    \item \textbf{Proof of Finite Differences converging to directional Jacobian sensitivities} \eqref{eq:directional_limit_clean}

\begin{proof}
Since \(f_\theta:\mathbb{R}^d \to \mathbb{R}^m\) is locally Fr\'echet differentiable at \(x\), there exists a linear map \(J_{f_\theta}(x)\in\mathbb{R}^{m\times d}\) such that
\begin{equation}
f_\theta(x+h u)
=
f_\theta(x)
+
J_{f_\theta}(x)(h u)
+
r(h),
\label{eq:frechet_expand_app}
\end{equation}
where the remainder satisfies
\begin{equation}
\frac{\|r(h)\|_2}{|h|} \to 0
\qquad \text{as } h \to 0.
\label{eq:frechet_remainder_app}
\end{equation}
Dividing \eqref{eq:frechet_expand_app} by \(h\) gives
\begin{equation}
\frac{f_\theta(x+h u)-f_\theta(x)}{h}
=
J_{f_\theta}(x)u
+
\frac{r(h)}{h}.
\label{eq:quotient_expand_app}
\end{equation}
By \eqref{eq:frechet_remainder_app}, we have
\[
\left\|
\frac{r(h)}{h}
\right\|_2 \to 0
\qquad \text{as } h\to 0.
\]
Therefore,
\[
\frac{f_\theta(x+h u)-f_\theta(x)}{h}
\to
J_{f_\theta}(x)u
\qquad \text{as } h\to 0.
\]
Taking squared \(\ell_2\)-norms on both sides and using continuity of the norm yields
\[
\lim_{h\to 0}
\left\|
\frac{f_\theta(x+h u)-f_\theta(x)}{h}
\right\|_2^2
=
\|J_{f_\theta}(x)u\|_2^2.
\]
This proves the claim.
\end{proof}

\item \textbf{Isotropic averaging of Directional Jacobian norm yields Frobenius Jacobian energy }\eqref{eq:isotropic_average_clean}

\begin{proof}
Let \(J := J_{f_\theta}(x)\in\mathbb{R}^{m\times d}\). Then
\begin{equation}
\|Ju\|_2^2
=
u^\top J^\top J u.
\label{eq:quad_form_app}
\end{equation}
Taking expectation over \(u\sim \mathrm{Unif}(\mathbb{S}^{d-1})\), we obtain
\begin{equation}
\mathbb{E}_u[\|Ju\|_2^2]
=
\mathbb{E}_u[u^\top J^\top J u].
\label{eq:expectation_quad_form_app}
\end{equation}
Using the trace identity \(v^\top A v = \operatorname{Tr}(A v v^\top)\), we have
\begin{equation}
\mathbb{E}_u[u^\top J^\top J u]
=
\mathbb{E}_u\!\left[\operatorname{Tr}(J^\top J\, u u^\top)\right]
=
\operatorname{Tr}\!\left(J^\top J\, \mathbb{E}_u[u u^\top]\right).
\label{eq:trace_identity_app}
\end{equation}
Since \(u\) is uniformly distributed on the unit sphere, its distribution is isotropic, and hence
\begin{equation}
\mathbb{E}_u[u u^\top] = \frac{1}{d} I_d.
\label{eq:sphere_isotropy_app}
\end{equation}
Substituting \eqref{eq:sphere_isotropy_app} into \eqref{eq:trace_identity_app} yields
\[
\mathbb{E}_u[\|Ju\|_2^2]
=
\operatorname{Tr}\!\left(J^\top J \cdot \frac{1}{d}I_d\right)
=
\frac{1}{d}\operatorname{Tr}(J^\top J).
\]
Finally, by the definition of the Frobenius norm,
\[
\operatorname{Tr}(J^\top J) = \|J\|_F^2.
\]
Therefore,
\[
\mathbb{E}_u[\|J_{f_\theta}(x)u\|_2^2]
=
\frac{1}{d}\|J_{f_\theta}(x)\|_F^2,
\]
as claimed.
\end{proof}
\end{enumerate}

\section{Experimental Details}\label{app:experimental_details}
We use the open-source DreamerV3 implementation, which is released under the MIT license, and the DeepMind Control Suite, which is released under the Apache-2.0 license.

Unless otherwise stated, GPLD uses posterior regularization with \(\lambda^{\mathrm{post}}_0=0.5\), no prior regularization, the square-root decay schedule from Algorithm~\ref{alg:lambda_decay}, imagination horizon \(H=25\), sampling fraction \(\rho=0.5\), decay scale \(c=1000\), and minimum coefficient \(\lambda_{\min}=0.001\). We use the 12M-parameter DreamerV3 configuration and keep all non-GPLD hyperparameters fixed between the baseline and GPLD variants.

Our primary metric is sample efficiency, measured by evaluation return as a function of environment steps. Learning curves report mean episodic return across available seeds, with shaded regions indicating one standard deviation around the mean; unless otherwise noted, experiments use five seeds. For aggregate curves, we report both raw and normalized aggregate return. In the normalized aggregate, each task is scaled by its final DreamerV3 baseline score before averaging, so that tasks contribute equally relative to their baseline performance.

\begin{table}[t]
  \caption{Training and evaluation hyperparameters. All non-GPLD hyperparameters are shared between DreamerV3 and GPLD-DreamerV3. Values are taken from the 12M DMC configuration, with GPLD-specific overrides listed separately.}
  \label{tab:training_hyperparameters}
  \centering
  \begin{tabular}{p{0.36\linewidth}p{0.25\linewidth}p{0.27\linewidth}}
    \toprule
    Hyperparameter & DreamerV3 & GPLD-DreamerV3 \\
    \midrule
    World-model size & 12M & 12M \\
    Benchmark & DeepMind Control & DeepMind Control \\
    % Proprioceptive observations & enabled & enabled \\
    % Pixel observations & secondary setting & secondary setting \\
    Environment action repeat & 2 & 2 \\
    Image resolution & \(64\times64\) & \(64\times64\) \\
    Number of parallel environments & 16 & 16 \\
    Training steps & \(10^6\) & \(10^6\) \\
    Train ratio, proprioceptive & 512 & 512 \\
    Train ratio, pixel & 512 & 512 \\
    Batch size & 16 & 16 \\
    Batch length & 64 & 64 \\
    Replay capacity & \(5\times10^6\) & \(5\times10^6\) \\
    Evaluation frequency & 5000 env. steps & 5000 env. steps \\
    Evaluation episodes during training & 3 & 3 \\
    Optimizer learning rate & \(4\times10^{-5}\) & \(4\times10^{-5}\) \\
    Optimizer \(\beta_1,\beta_2\) & \(0.9,0.999\) & \(0.9,0.999\) \\
    Optimizer epsilon & \(10^{-20}\) & \(10^{-20}\) \\
    Weight decay & 0 & 0 \\
    AGC coefficient & 0.3 & 0.3 \\
    RSSM deterministic size & 2048 & 2048 \\
    RSSM stochastic variables & 32 & 32 \\
    RSSM classes per variable & 16 & 16 \\
    RSSM hidden size & 256 & 256 \\
    Network depth & 16 & 16 \\
    Network units & 256 & 256 \\
    Loss scales \((\beta_{\mathrm{pred}},\beta_{\mathrm{dyn}},\beta_{\mathrm{rep}})\) & \((1.0,1.0,0.1)\) & \((1.0,1.0,0.1)\) \\
    Imagination horizon & 25 & 25 \\
    \(\lambda^{\mathrm{post}}_0\) & 0 & 0.5 \\
    \(\lambda^{\mathrm{prior}}_0\) & 0 & 0 \\
    Sampling fraction \(\rho\) & --- & 0.5 \\
    Penalty decay & --- & square-root decay \\
    Decay scale \(c\) & --- & 1000 \\
    Minimum coefficient \(\lambda_{\min}\) & --- & 0.001 \\
    \bottomrule
  \end{tabular}
\end{table}

Table~\ref{tab:training_hyperparameters} lists the main training and evaluation hyperparameters used in our experiments. DreamerV3 and GPLD-DreamerV3 share all non-GPLD hyperparameters; the only differences are the posterior gradient penalty, sampling fraction, and decay schedule.

%%%%%%%%%%%%%%%%%%%%%%%%%%%%%%%%%%%%%%%%%
% \include{sections/sec99C_Results}
%%%%%%%%%%%%%%%%%%%%%%%%%%%%%%%%%%%%%%%%%
\section{DMC Proprioceptive Results}\label{app:proprio_2M}
\begin{figure}[ht]
    \centering
    \includegraphics[width=1.0\linewidth]{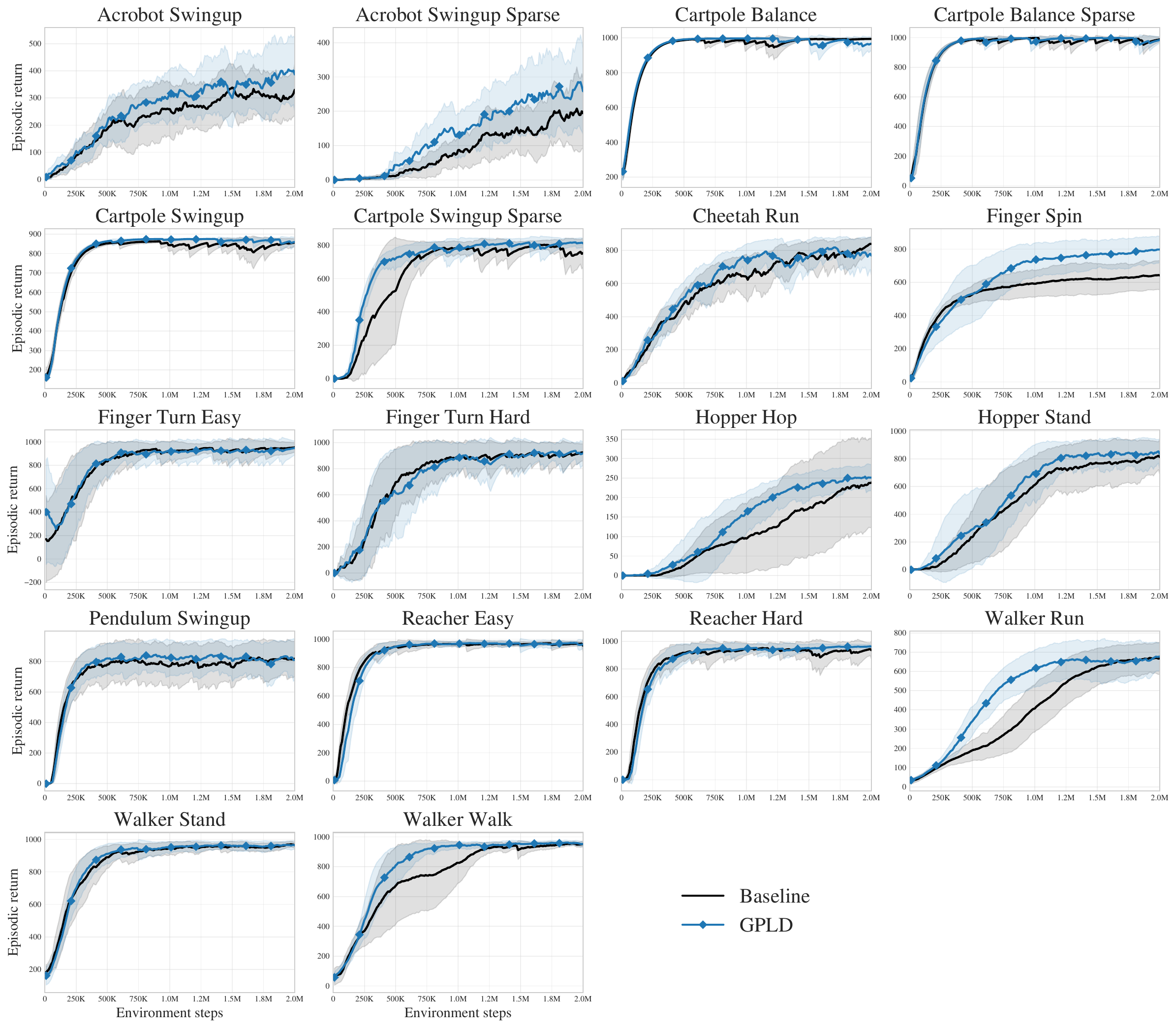}
    \caption{DMC Proprioceptive Individual Tasks}
    \label{fig:proprio_2M_grid}
\end{figure}

Fig \ref{fig:proprio_2M_grid} shows that the early GPLD gains generally persist or remain competitive over longer horizons, while quadruped tasks are separated in the main text because their benefits emerge most clearly only beyond the 1M-step regime.

\begin{table}[t]
  \caption{Full 1M-step proprioceptive DMC results. Scores are reported as mean $\pm$ standard deviation across seeds, using checkpoint evaluation with 20 episodes per seed. GPLD-DreamerV3 improves the normalized aggregate mean by \(17.7\%\).}
  \label{tab:app_proprio_1m_all_tasks}
  \centering
  \small
  \setlength{\tabcolsep}{5pt}
  \begin{tabular}{lccc}
    \toprule
    Task & DreamerV3 & GPLD-DreamerV3 & Relative gain \\
    \midrule
    acrobot-swingup & \(238.1 \pm 15.5\) & \(304.6 \pm 15.4\) & \(+27.9\%\) \\
    acrobot-swingup-sparse & \(99.7 \pm 18.0\) & \(135.6 \pm 58.4\) & \(+36.0\%\) \\
    cartpole-balance & \(990.0 \pm 6.6\) & \(997.1 \pm 0.0\) & \(+0.7\%\) \\
    cartpole-balance-sparse & \(999.9 \pm 0.1\) & \(1000.0 \pm 0.0\) & \(+0.0\%\) \\
    cartpole-swingup & \(860.3 \pm 0.6\) & \(870.3 \pm 4.1\) & \(+1.2\%\) \\
    cartpole-swingup-sparse & \(729.3 \pm 57.1\) & \(816.2 \pm 29.2\) & \(+11.9\%\) \\
    cheetah-run & \(672.6 \pm 15.3\) & \(762.8 \pm 39.0\) & \(+13.4\%\) \\
    finger-spin & \(598.2 \pm 79.4\) & \(735.2 \pm 97.3\) & \(+22.9\%\) \\
    finger-turn-easy & \(941.4 \pm 8.6\) & \(903.2 \pm 39.6\) & \(-4.1\%\) \\
    finger-turn-hard & \(853.0 \pm 75.8\) & \(882.5 \pm 55.1\) & \(+3.5\%\) \\
    hopper-hop & \(81.8 \pm 79.0\) & \(197.4 \pm 41.7\) & \(+141.4\%\) \\
    hopper-stand & \(683.6 \pm 156.0\) & \(730.5 \pm 125.0\) & \(+6.9\%\) \\
    pendulum-swingup & \(760.7 \pm 121.3\) & \(821.2 \pm 18.4\) & \(+8.0\%\) \\
    reacher-easy & \(947.1 \pm 28.5\) & \(968.1 \pm 8.6\) & \(+2.2\%\) \\
    reacher-hard & \(943.9 \pm 30.5\) & \(947.9 \pm 23.9\) & \(+0.4\%\) \\
    walker-run & \(458.5 \pm 122.8\) & \(627.1 \pm 77.2\) & \(+36.8\%\) \\
    walker-stand & \(942.8 \pm 16.1\) & \(957.8 \pm 8.5\) & \(+1.6\%\) \\
    walker-walk & \(878.3 \pm 86.9\) & \(946.2 \pm 12.7\) & \(+7.7\%\) \\
    % \midrule
    % Raw aggregate mean & \(704.4\) & \(755.8\) & \(+7.3\%\) \\
    % Normalized aggregate mean & \(1.000\) & \(1.177\) & \(+17.7\%\) \\
    \bottomrule
  \end{tabular}
\end{table}

\section{DMC 
Pixel Results}
\subsection{Individual learning plots}\label{app:vision_grid}
\begin{figure}[ht]
    \centering
    \includegraphics[width=1.0\linewidth]{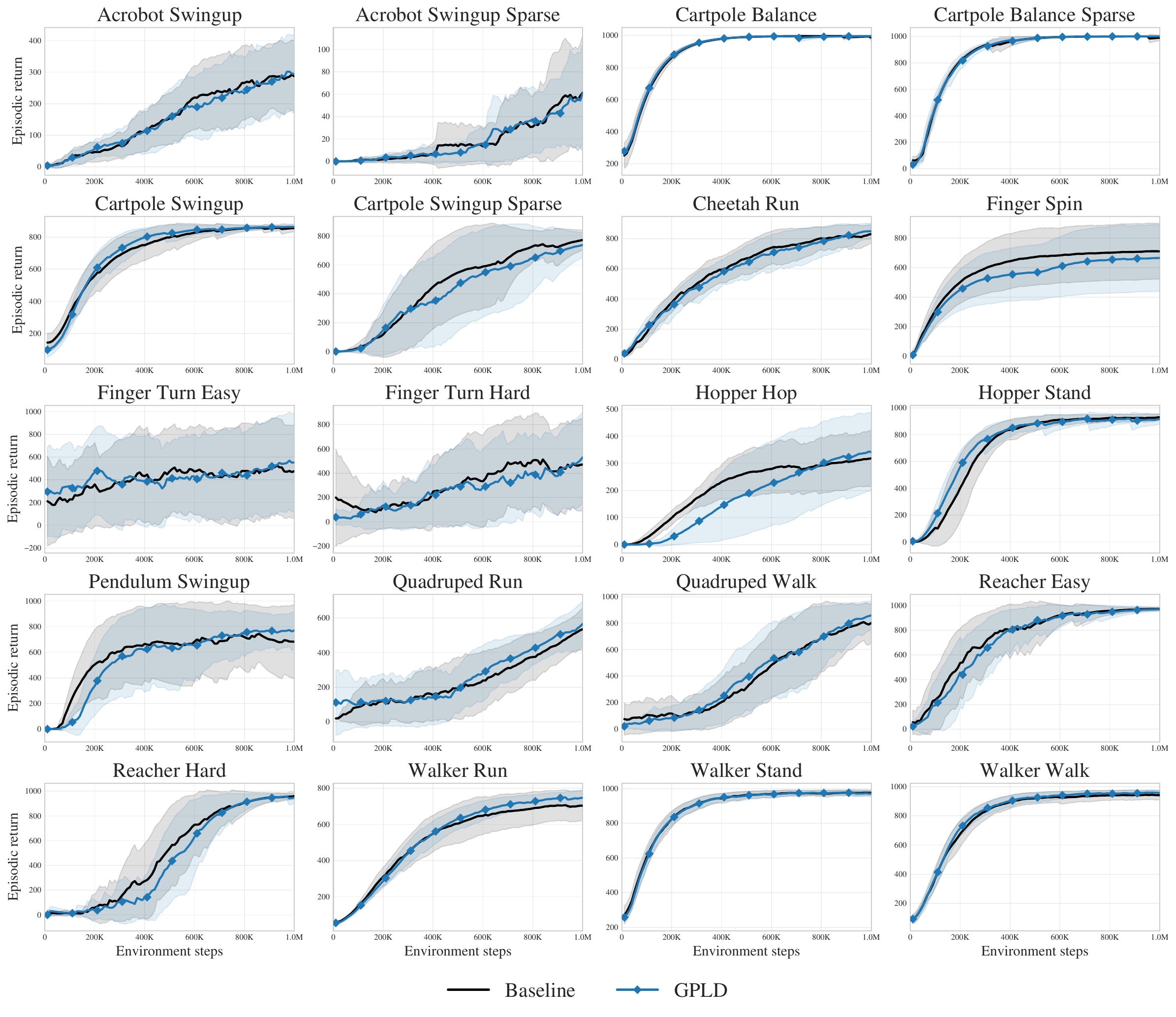}
    \caption{DMC Pixel Individual Tasks}
    \label{fig:vision_1M_grid}
\end{figure}

Fig. \ref{fig:vision_1M_grid} shows individual plots for the DMC Vision tasks. 

\subsection{Encoder-decoder warm-start diagnostic for pixel observations}
\label{app:vision_warmstart}

To better understand why GPLD yields weaker aggregate gains under pixel observations, we run an encoder-decoder warm-start diagnostic. In this setting, the encoder and decoder are initialized from a previously trained DreamerV3 baseline run, while all other components are reinitialized. The encoder and decoder are not frozen; they continue to train during the new run. Both DreamerV3 and GPLD-DreamerV3 use the same encoder-decoder initialization, so differences in performance reflect how the methods use the improved visual initialization rather than a different starting representation.

\begin{figure*}[ht]
    \centering
    \begin{subfigure}[t]{0.49\linewidth}
        \centering
        \includegraphics[width=\linewidth]{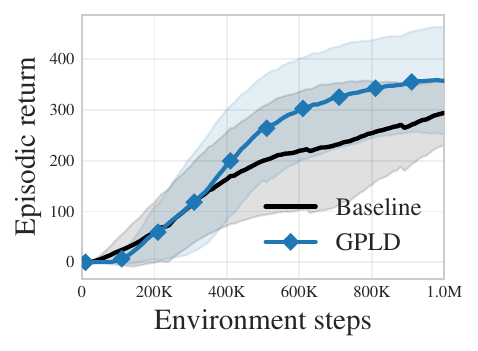}
        \caption{Hopper-hop}
    \end{subfigure}
    \hfill
    \begin{subfigure}[t]{0.49\linewidth}
        \centering
        \includegraphics[width=\linewidth]{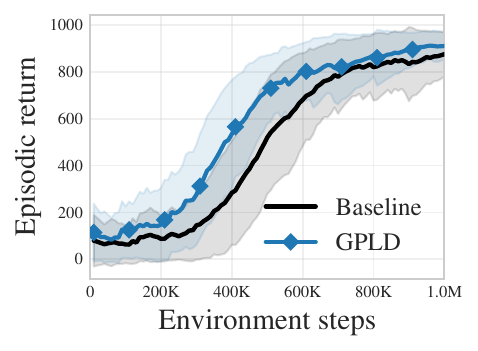}
        \caption{Quadruped-walk}
    \end{subfigure}
    \caption{Encoder-decoder warm-start diagnostic for pixel observations. The encoder and decoder are initialized from a trained DreamerV3 baseline run and continue training, while all other components are reinitialized. With the same visual warm start, GPLD-DreamerV3 improves performance on hopper-hop and quadruped-walk, suggesting that visual representation learning may partially mask the effect of latent-dynamics regularization in fully end-to-end pixel training.}
    \label{fig:vision_warmstart}
\end{figure*}

Figure~\ref{fig:vision_warmstart} shows results on hopper-hop and quadruped-walk. Under the same warm start, GPLD improves performance over DreamerV3 on these tasks. This supports the interpretation that the weaker fully end-to-end pixel gains may partly arise because visual representation learning and latent-dynamics learning must be solved simultaneously. We treat this experiment as a diagnostic rather than a main empirical claim.

\newpage

\section{Computational Cost Analysis}\label{app:wall_clock}
Computation cost analysis: (Avg Baseline time = 2:01:29/seed for 500k env steps.)
\begin{table}[t]
  \caption{Computation-time overhead of GPLD relative to vanilla DreamerV3 across different $(\rho,\lambda^{post},\lambda^{prior})$ settings.  Runtime is reported as the mean ratio over runs, with standard deviation.}
  \label{tab:compute_overhead}
  \centering
  \begin{tabular}{rrrrr}
    \toprule
    $\rho$ & $\lambda^{post}$ & $\lambda^{prior}$ & Mean $\pm$ Std (ratio) & Diff (s) \\
    \midrule
    0.25 & 0    & 0.5  & $1.01 \pm 0.022$ & 74.40 \\
    0.25 & 0.5  & 0    & $1.11 \pm 0.063$ & 770.10 \\
    0.5  & 0    & 0.5  & $1.15 \pm 0.032$ & 1108.00 \\
    \textbf{0.5} & \textbf{0.25} & \textbf{0} & $\textbf{1.23} \pm \textbf{0.028}$ & \textbf{1668.00} \\
    0.5  & 0.25 & 0.25 & $1.44 \pm 0.027$ & 3200.60 \\
    \textbf{0.5} & \textbf{0.5} & \textbf{0} & $\textbf{1.51} \pm \textbf{0.157}$ & \textbf{3712.10} \\
    0.5  & 0.5  & 0.5  & $1.44 \pm 0.027$ & 3177.40 \\
    0.5  & 0.75 & 0.75 & $1.43 \pm 0.037$ & 3118.40 \\
    0.75 & 0    & 0.5  & $1.26 \pm 0.026$ & 1927.30 \\
    0.75 & 0.5  & 0    & $1.46 \pm 0.059$ & 3381.80 \\
    1.0  & 0    & 0.5  & $1.40 \pm 0.029$ & 2895.20 \\
    1.0  & 0.5  & 0    & $1.68 \pm 0.100$ & 4920.40 \\
    \bottomrule
  \end{tabular}
\end{table}

GPLD introduces additional computation because the gradient penalty requires vector-Jacobian products. The main algorithmic cost driver is the sampling fraction \(\rho\), which determines the fraction of batch states on which the penalty is evaluated. In contrast, changing the penalty coefficient \(\lambda^{\mathrm{post}}\) does not substantially change the number of derivative evaluations and should not, by itself, determine the algorithmic overhead.

Table~\ref{tab:compute_overhead} reports wall-clock overhead relative to vanilla DreamerV3. For the default setting used in our main experiments, \(\lambda^{\mathrm{post}}_0=0.5\) and \(\rho=0.5\), the measured runtime ratio is \(1.51\). However, this value should be interpreted cautiously: the standard deviation is relatively large, and a nearby posterior-only run with the same sampling fraction, \(\lambda^{\mathrm{post}}_0=0.25\) and \(\rho=0.5\), has a substantially lower runtime ratio of \(1.23\). Since these two settings require the same type of gradient-penalty computation, the difference suggests that the \(1.51\) measurement likely reflects wall-clock variability in the execution environment in addition to the algorithmic overhead of GPLD.

Overall, the timing results indicate that GPLD adds measurable overhead relative to DreamerV3, but the \(1.51\times\) value should be viewed as a conservative high-end wall-clock estimate rather than the inherent cost of the method. Runs with the same sampling fraction suggest a more typical overhead closer to the \(1.2\)--\(1.3\times\) range, although precise runtime depends on hardware utilization and system-level variability.

\newpage

\section{Local sensitivity analysis}
\label{app:local_sensitivity}
To verify that GPLD changes the learned world model in the intended direction, we estimate the local sensitivity of the posterior and prior distributions during training. For each saved checkpoint, we generate trajectories and perturb the corresponding latent inputs. For the posterior, we perturb the encoder-dependent input \(e_t\); for the prior, we perturb the deterministic state \(h_t\). We then measure the change in the output distribution using KL divergence and normalize by the perturbation norm.

Let \(r_\phi(\cdot\mid u_t)\) denote either the posterior or prior distribution evaluated at its corresponding input \(u_t\). For a perturbation \(\delta\), we estimate
\[
\frac{
D_{\mathrm{KL}}\!\left(r_\phi(\cdot\mid u_t)\,\|\,r_\phi(\cdot\mid u_t+\delta)\right)
}{
\|\delta\|_2
},
\]
with perturbation magnitudes ranging from \(0.1\%\) to \(10\%\) of the average input norm. We average this quantity across time steps, perturbation magnitudes, and seeds. Lower values indicate that the learned latent distribution changes less under small input perturbations.

\begin{figure}[t]
  \centering
  \begin{minipage}[t]{0.24\linewidth}
    \centering
    \includegraphics[width=\linewidth]{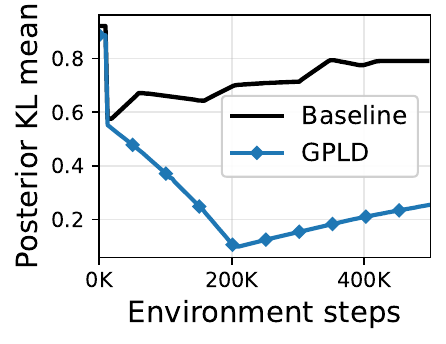}
    {\footnotesize (a) Walker walk: Posterior aggregate}
  \end{minipage}\hfill
  \begin{minipage}[t]{0.24\linewidth}
    \centering
    \includegraphics[width=\linewidth]{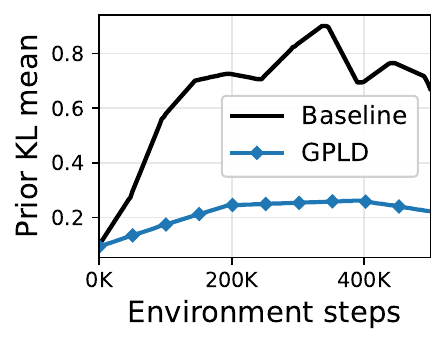}
    {\footnotesize (b) Walker walk: Prior aggregate}
  \end{minipage}\hfill
  \begin{minipage}[t]{0.24\linewidth}
    \centering
    \includegraphics[width=\linewidth]{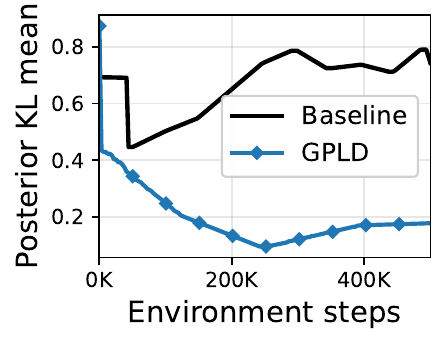}
    {\footnotesize (c)Hopper stand: posterior aggregate}
  \end{minipage}\hfill
  \begin{minipage}[t]{0.24\linewidth}
    \centering
    \includegraphics[width=\linewidth]{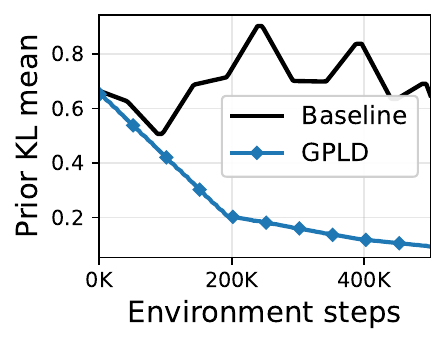}
    {\footnotesize (d) Hopper stand: prior aggregate}
  \end{minipage}
  \caption{Aggregate local sensitivity of the learned posterior and prior distributions for Walker walk and Hopper stand environments. We perturb the corresponding latent inputs and measure the KL change in the output distribution normalized by the perturbation norm. GPLD reduces posterior sensitivity relative to DreamerV3, consistent with the intended effect of the gradient penalty.}
  \label{fig5.5:lipschitz_analysis}
\end{figure}

Figure~\ref{fig5.5:lipschitz_analysis} shows aggregate local sensitivity across tasks. GPLD reduces the sensitivity of the learned posterior distribution relative to DreamerV3, consistent with the intended effect of the gradient penalty. The prior sensitivity is also reduced in several cases, even though GPLD is applied directly only to the posterior, suggesting that the posterior regularizer influences the prior through the DreamerV3 dynamics loss.

%%%%%%%%%%%%%%%%%%%%%%%%%%%%%%%%%%%%%%%%%%%%%%%%%%%%%%%%%%%%

\newpage

\end{document}